\documentclass[conference,compsoc]{IEEEtran}
\usepackage{float}
\usepackage{booktabs}
\usepackage{graphicx,times,amsmath,amssymb,multirow,subcaption,color}
\usepackage{mathrsfs}
\usepackage[justification = centering]{caption}
\usepackage[LGR,T1]{fontenc}
\usepackage[latin9]{inputenc}
\usepackage{wrapfig}
\usepackage{array,textcomp,stackrel,url,mathtools,enumerate}
\usepackage{multirow} 
\usepackage{multicol} 
\usepackage{booktabs,helvet,courier,float,csquotes}
\usepackage{algorithm,algcompatible}
\usepackage{flushend}
\usepackage{csquotes}
\usepackage[colorinlistoftodos]{todonotes}
\usepackage{color}
\usepackage{color,soul}
\usepackage{blindtext}
\usepackage[inline]{enumitem}
\usepackage{pdfpages}
\usepackage{pdflscape}
\usepackage{wrapfig}
\usepackage{cite}
\newtheorem{definition}{Definition}
\newtheorem{specification}{Specification}

\begin{document}

\title{Lifelong Testing of Smart Autonomous Systems by Shepherding a Swarm of Watchdog Artificial Intelligence Agents}
\author{
\IEEEauthorblockN{Hussein Abbass}
\IEEEauthorblockA{\textit{School of Engineering \& IT} \\
\textit{University of New South Wales}\\
Canberra, Australia\\
h.abbass@unsw.edu.au}
\and
\IEEEauthorblockN{John Harvey}
\IEEEauthorblockA{\textit{School of Engineering \& IT} \\
\textit{University of New South Wales}\\
Canberra, Australia\\
johnpaulharvey1@me.com}
\and
\IEEEauthorblockN{Kate Yaxley}
\IEEEauthorblockA{\textit{School of Engineering \& IT} \\
\textit{University of New South Wales}\\
Canberra, Australia\\
k.yaxley@unsw.edu.au}
        }

\markboth{AI Watchdog}
{}
\maketitle

\begin{abstract}     
Artificial Intelligence (AI) technologies could be broadly categorised into Analytics and Autonomy. Analytics focuses on algorithms offering perception, comprehension, and projection of knowledge gleaned from sensorial data. Autonomy revolves around decision making, and influencing and shaping the environment through action production. A smart autonomous system (SAS) combines analytics and autonomy to understand, learn, decide and act autonomously. To be useful, SAS must be trusted and that requires testing. Lifelong learning of a SAS compounds the testing process. In the remote chance that it is possible to fully test and certify the system pre-release, which is theoretically an undecidable problem, it is near impossible to predict the future behaviours that these systems, alone or collectively, will exhibit.  While it may be feasible to severely restrict such systems\textquoteright \ learning abilities to limit the potential unpredictability of their behaviours, an undesirable consequence may be severely limiting their utility. In this paper, we propose the architecture for a watchdog AI (WAI) agent dedicated to lifelong functional testing of SAS. We further propose system specifications including a level of abstraction whereby humans shepherd a swarm of WAI agents to oversee an ecosystem made of humans and SAS. The discussion extends to the challenges, pros, and cons of the proposed concept.
\end{abstract}

\begin{IEEEkeywords}
Artificial Intelligence Testing, Autonomous Systems Certification, Swarm Intelligence, Smart Safety Nets, Watchdog Artificial Intelligence
\end{IEEEkeywords}

\section{Introduction}

The philosophical underpinnings of testing is to discover problematic states where a system\textquoteright s action is unacceptable. Classically, testing algorithms and protocols such as formal methods~\cite{boyer1983proof,clarke2018model} are used to check if specifications are fulfilled, while user acceptance are used to define what is `right' during the validation process~\cite{bianchi2016testing}. Some literature attempts to tackle some complex problems of verification methods failure~\cite{brat2005challenges}. For example, under very restrictive Markovian assumptions, Arora \& Rao~\cite{arora2016probabilistic} attempted to verify incomplete models. Scenarios to generate the requirements for verification have been used to approximate the requirement space ahead of design and release~\cite{heitmeyer2015obtaining}. The research is mostly premised on the assumption that testing is right-skewed towards the pre-product release stage and/or that complete system specifications are known in advance~\cite{helle2016testing}. These assumptions are inadequate for the `smartness' dimension of a smart autonomous system (SAS).

Latest attempts to develop testing methods for SAS unfolded two research directions, using machine learning for testing and testing for machine learning~\cite{nelson2004survey,pulina2010abstraction}. Machine learning methods have been used at run-time for verification~\cite{pulina2010abstraction} but the Probably Approximate Correct (PAC) bounds~\cite{vapnik2006estimation} are unrealistic settings for most practical AI systems~\cite{russell2015research}. Preliminary attempts to verify the learning algorithms are in their infancy, let alone validating and testing these algorithms, and \textquote[{\cite[p. 108]{russell2015research}}]{\textit{much remains to be done before it will be possible to have high confidence that a learning agent will learn to satisfy its design criteria in realistic contexts.}} More challenges exist when SASs operate in a socio-technical setting~\cite{endsley2017here,russell2015research}.

As we transform the nature of machine decision-making from responding to scripted triggers with fixed logic and little smartness to `smarter' responses that learn from the interaction, change their form and internal control logic, and adapt to the contexts they get situated within, SASs display different characteristics from classic software systems. 

This is primarily due to two fundamental characteristics. First, a smart system continues to learn and could change the control logic generating its actions. This change is driven by interactions with the environment; thus, its control logic is not fixed and could not be completely defined in-advance; at least, this is the case for future complex SAS. This limits the use of structural testing methodologies although it is possible to borrow and adapt ideas from automatic data generation methods such as those presented in~\cite{gotlieb1998automatic}.

Second, a reasonably sophisticated SAS usually relies either on distributed services or very complex system-of-systems design. The practical implications is that a SAS relies on other proprietary sub-systems that makes the availability of the code or access of internal states of these proprietary sub-systems infeasible. Thus, certification of the system as a whole faces more challenges than those faced by complex software systems (eg the software system onboard of an aircraft). Blackbox testing using functional testing is also problematic because in the absence of a complete representation of the internal states of a SAS, the mathematical assumptions that the SAS is acting as a functional mapping breaks down; that is, the same set of inputs could generate different outputs based on the hidden internal states of SAS. 

To unfold the challenges facing testing for SAS, three dimensions are worth differentiating: automation, autonomy and the learning agent. Lee and See~\cite{lee2004trust} define automation as a \textquote[{\cite[p. 50]{lee2004trust}}]{\textit{technology that actively selects data, transforms information, makes decisions, or controls processes}}; automation is about the technological capacity (ie skills and competencies) of an entity to perform a task. Autonomy, is concerned with the opportunity afforded to the system to act~\cite{abbass2016trusted}. The difference between the capacity and opportunity is a primary source of risk. A SAS acting without the capacity to perform the task will make mistakes. Denying a SAS to act, when it safely can, is inefficiency. The learning agent not only improves SAS\textquoteright \ capacity (i.e. automation), but a truly smart and self-aware agent needs to work on reducing this difference. Self-awareness is necessary for the agent to be able to assess its own capacity (automation) to adjust its level of autonomy.

Here lies two fundamental research challenges for testing of SAS: the behavioural space (internal control logic and associated actions) of SAS is not fixed, and defining what behaviour is  `right' may change from one operational context to another. 
These challenges are compounded with commercial pressures to expedite the production of SAS, causing shorter testing cycles and hidden risks which, when they eventually surface, may lead to significant negative consequences~\cite{augusto2011living}. Moreover, the coupling between the physical and cyber layers of the system and the system\textquoteright s interface with humans~\cite{romero2016operator} leads to a level of complexity that severely limits the efficacy of any segregated testing approaches. These challenges call for new ways of thinking about the testing methodologies that could become more practical for a complex SAS.

In the animal behaviour testing domain, researchers adopt a different approach, using standardised experimental contexts, pioneered by researchers such as Serpell and Hsu~\cite{serpell2001development}. Within the boundaries of these experimental contexts, stimuli are triggered to provoke a behaviour. The set of behaviours in response to the stimuli are then compared statistically against other animals that were exposed to the same contexts. The animal under testing is then assigned to the most appropriate group. Animal behavioural testing is used either to evaluate the expressed behaviour of an animal in response to stimuli or as a mean to categorise animals into behavioural categories~\cite{diederich2006behavioural}.

This dynamic profiling approach to testing is designed out of necessity because of the unbounded behavioural space of animals and the need for flexibility to accommodate novel behaviours. The behaviour of one animal could be categorised as unacceptable (eg wild) or acceptable (eg friendly) based on the profile of other animal groups. SAS can, and are very likely to, get smarter than animals, their behavioural space is far more complex than the discrete categories an animal profile may fall into and the objective of their testing is not to merely categorise them, but to regulate them. This data-driven testing approach needs to operate on the functional level, without access to the internal logic, code, or states of the agent.

In this paper, we propose artificial intelligence (AI) watchdogs (WAIs)\footnote{The abbreviation WAI is also the customary greeting in Thailand. It signals safety by having both hands clasped together in front of someone to demonstrate there are not unsafe holding of objects like a weapon. This inspired the abbreviation where a WAI agent welcomes a SAS to operate safely.}. These WAIs need to assume the role of the human expert in designing standardised experimental contexts autonomously while learning these contexts on the fly. Each WAI is an intelligent testing agent with a portfolio of responsibilities and falls into two categories: Behaviour Smart Safety Net (BSSN) WAI agents, and shepherding WAI control agents. The BSSN WAI agents oversee a SAS for performance assurance and categorise SAS behaviours, while the shepherd WAI agents oversee and control the BSSN agents.

The remainder of the paper is structured as follows. The conceptual framework of WAIs is presented in Section~\ref{ConceptualFramework}, followed by a mathematical formulation of the tasks to be performed by WAIs in Section~\ref{Formulation}. We then present the main specifications for the design of WAIs, followed by a discussion and associated challenges of WAIs in Section~\ref{Discussion} then conclusion and future work in Section~\ref{Conclusion}.

\section{Conceptual framework}\label{ConceptualFramework}

\begin{figure}[hptb]
\centering 
 \includegraphics[width=0.95\columnwidth]{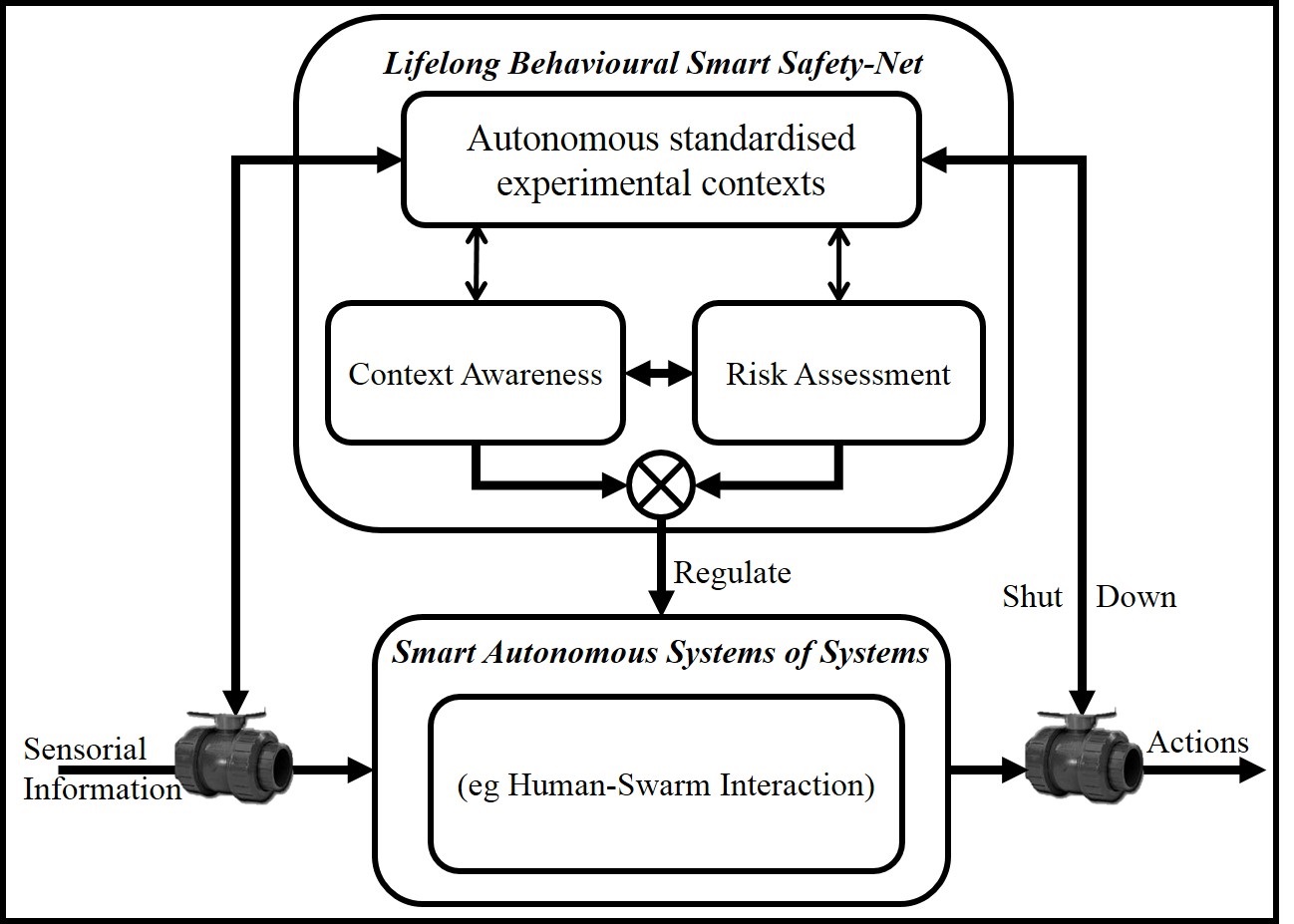}
 \caption{The conceptual framework of the lifelong BSSN WAI agent and its interaction with SASs.}\label{fig:bssn1}
\end{figure}

Figure~\ref{fig:bssn1} presents the overall conceptual framework, where BSSN WAI agents oversee the autonomous system, or the autonomous system-of-systems, to govern the inflow and outflow of information to and from the system, respectively. It can interrogate the system to test its behaviour by sending its own input signals and listen to the output. BSSN WAIs can sit silent watching and listening to the sensors and actuators, and the corresponding information passing through the input and output control gates to risk-assess the system\textquoteright s courses of actions and their consequences. WAIs can equally prevent the SAS from acting by shutting down the output control gates. Preventing unsafe action is a responsibility that WAI needs to carry because it is an important action in its own right.
%
%

The success of BSSN revolves around its ability to be context-aware to understand the context SAS is embedded within and evaluate consequences of SAS\textquoteright \ decisions. It is not the responsibility of BSSN to be smarter than the SAS in making the best decision for a given context - this would obviate the need for the SAS - nor is it the responsibility of the BSSN to produce decisions or decide what the best decision is in a given context. Rather, it is the role of the BSSN to ensure that the outputs of the SAS remain within acceptable bounds of behaviour. In a nutshell, BSSN is designed to be a \textquoteleft conservative\textquoteright , but adaptive, autonomous watchdogs for SAS, making sure the SAS operates within acceptable bounds of behaviour.

A BSSN agent has three components. One is responsible for contextual awareness, whereby analytics are used to derive patterns from sensorial information and fuse the information into higher-level knowledge. The second is responsible for risk assessment to evaluate the context and actions of SAS to ensure safety and that it conforms to the behavioural constraints on its action set. The third component is responsible for autonomous standardised experimental contexts. Given the contexts the SAS are faced with, it dynamically generates scenarios (ie a semantically coupled dataset) to test SAS, collects the behavioural responses from the system, profiles the behaviours, and evaluates the suitability of SAS for these contexts. It is important to emphasise that the software nature of an autonomous system allows the WAI agents to ask what-if questions of the SAS; thus, actions under standardised experimental contexts scenarios do not proceed to actuators; instead, they proceed to the WAI agent for evaluation. In simple terms, when a WAI agent is conducting a standardised experimental context test, SAS is just thinking aloud, not acting on the environment it is embedded within  - i.e. it is disengaged from the real-world.

\subsection{Design Requirements}

WAI needs to fulfil a number of design requirements listed below: 
\begin{enumerate}

\item It needs to be verifiable while being adaptable at the same time. Verifiability is key for safety assurance of the system. Adaptability is vital to survive potentially extreme dynamics and changes in its surrounding environment and the SAS it is testing. However, adaptability needs to be a very slow conservative process that does not break down the integrity of WAI.

\item It needs to be faster than SAS in its decision cycle to avoid time lags in SAS' response. Considering that the testing agent does not need to replicate what the autonomous system does, but instead it needs to assess the decisions produced by the autonomous system given a context, different layers of the architecture of the testing agent need to work on different time scales. The sub-system responsible for the output gate needs to be the fastest to avoid time lag in releasing actions out of SAS because this can be prohibitive in time-critical applications.
%

\item The decoupling of BSSN from SASs entails some functionalities in SAS will need to be duplicated in BSSN, especially those related to perception. However, this replication can be thought of independent of the way these functions are implemented within SAS. The possible cost associated with duplication of functions is balanced with the benefits of integrity assurance of SAS.

\end{enumerate}

The first requirement will be achieved in the WAI architecture by separating the design patterns to oversee action execution of the autonomous system from the processes that generate these design patterns and adapt them through action production within the testing agent. The second requirement  will be achieved by coupling symbolic (to achieve transparency) and non-symbolic (to achieve speed) knowledge representation within the testing agent. The third requirement  is guaranteed by design as we will not make any assumption on how SAS make decisions, neither will WAI requires access to the internal states of this SAS. Put simply, SAS will be a black-box to BSSN.

Decisions made by WAI to regulate the actions of SAS use declarative knowledge represented in the form of \enquote{If $\dots$ Then $\dots$} rules due to threefold of advantages. First, these production rules are needed for action production and reasoning. Second, their interpretability eases the way to certification and if needed, can change form to propositions suitable for propositional constraint satisfaction engines. However, this layer is similar to classic safety-nets; a firewall that inspects traffic using pattern matching techniques. Hence, the third advantage is that we could update the knowledge/constraints in this layer by means of machine learning by using either a rule-based representation similar to the one in~\cite{shafi2011evaluationdetection,shafi2009andetection} or a non-symbolic neural-based representation similar to the one in~\cite{dam2008neuralbasedsystems}.

\section{Formulation}\label{Formulation}

\subsection{SAS Testing Spaces}

Testing a system/agent is done relative to a set of user specifications. These specifications could be describing the physical behaviour of the robot such as the speed limits on an autonomous car; safety behaviour of the robot such as maximum speed and acceleration allowed in a school zone; or legal considerations such as the 
legality surrounding the autonomous car to operate in certain zones. Some of these specifications form hard constraints that violating any of them would deem a system unfit for purpose, while others are soft constraints where violating them could lead to a level of discomfort incurring a cost or penalty, but does not impact the fitness of the system for the purpose it was designed for. An example of a soft constraint is an autonomous car with a maximum desirable turning rate at a corner to avoid causing discomfort to its passengers, but in a case of emergency, this constraint could get violated.

The testing problem could get formulated either purely as a constraint satisfaction problem where only hard constraints are considered, or as an optimisation problem, where hard constraints need to be respected all the time while minimising the cost of violating soft constraints. Denoting hard constraints in a standard form by $g(v)\le0$, where $v$ is the behaviour parameters, and soft constraints by $h(v)\le0$, given a specific behaviour of a SAS, $\hat{v}$, the total hard constraints violation is denoted by $V_g(\hat{v})$, while total soft constraints violation is denoted by $V_h(\hat{v})$.

\begin{figure}[hpbt]
\centering 
 \includegraphics[width=0.95\columnwidth]{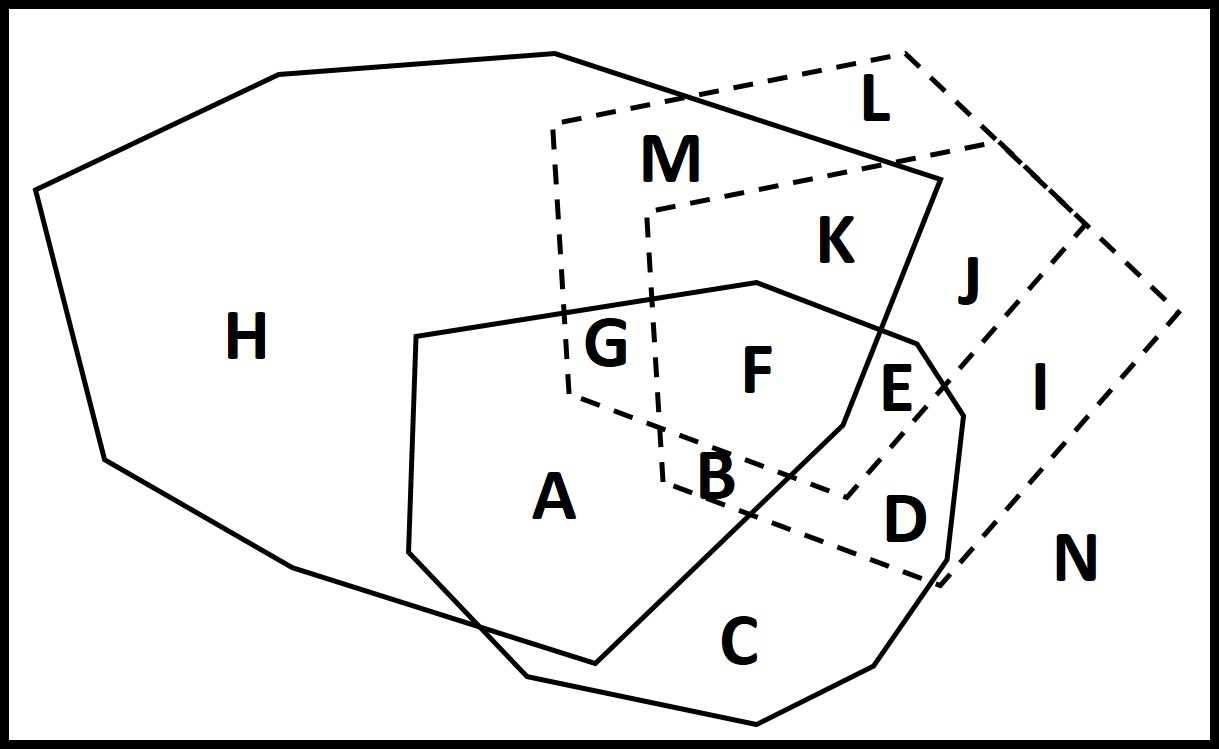}
 \caption{The Testing Space.}\label{fig:testing}
\end{figure}

Figure~\ref{fig:testing} presents a conceptual diagram of different behavioural sub-spaces presented below. For clarity, we assume all sub-spaces are polyhedron, with an alphabet associated with each polyhedron.

The space where all hard requirements/constraints are satisfied is denoted by $S_g = \cup(A,B,C,D,E,F,G)$, while the space where all soft requirements/constraints are satisfied is denoted by $S_h = \cup(A,B,G,F,K,M,H)$. The behaviour space of an Autonomous system before a learning cycle is denoted by $S_b = \cup(B,D,F,E,I,J,K)$, with this behaviour space shifting after learning to $S_a = \cup(E,F,G,J,K,L,M)$. 

In the absence of soft requirements, the aim of the testing problem is to ensure that a SAS\textquoteright \ set of actions are always within $S_g$. An alternative formulation of the testing problem is to learn/discover the space of hard constraint violations, which is, $S_{bhv} = \cup(I,J,K)$ for pre-learning behaviours and $S_{ahv} = \cup(J,K,L,M)$ for post-learning behaviours. After considering the soft-constraints, the unacceptable, including less desirable, behaviours for the pre-learning SAS expand to $S_{bv} = \cup(I,J,K,E,D)$ and for the post-learning SAS is $S_{ahv} = \cup(J,K,L,M,E)$.

This abstract example describes the complexity of testing a SAS. While one could assume that $S_g$ and $S_h$ are both known in advance and without loss of generality, we could assume them to form a bounded set such as the polyhedrons in our representation, the challenge in testing is to estimate $S_b$ and $S_a$. As the SAS continues to learn, its behaviour space continues to contract and expand the sub-spaces that were previously estimated. In our example, $\cup(J,K,E,F)$ is the common behaviour space before and after learning, while $\cup(B,D,I)$ were replaced with $\cup(G,M,L)$.

\subsection{BSSN WAI}

We will first assume that the SAS is in a state where its behavioural space is $S_{b}$. Learning will cause it to move to a different state that corresponds to space $S_{a}$, but first we will consider the former state. The highest priority for the WAI agents is to discover the areas $I,J$ and $K$ because the agent violates the hard constraints on its behavioural envelop. These could be for example violating an ethical or a legal constraint. The second priority is to discover $D$ and $E$, where the agent is fulfilling all hard constraints but not the soft constraints. The third priority is to discover areas $B$ and $F$, which represent the capacity of the agent to act right (level of automation).

After a learning cycle of SAS, WAI agents need to learn the change. In particular, they need to learn that the SAS no longer generates actions in areas $B,D$ and $I$. New behaviours that violate hard constraints are generated in areas $L$ and $M$. Behaviours that were acceptable but were then lost (possibly due to the forgetting phenomenon in learning models) are represented by area $D$, where the SAS was capable of generating behaviours that are acceptable, but then lost this capacity. Moreover, the WAIs need to learn that SAS has built an extra capacity to perform in area $G$.

For WAI agents to discover these areas, they need computational methods. We present two in particular below. We will denote a BSSN WAI $i$ by $\pi_i$. A $\pi_i$ agent has two tasks. The first is defined with the following problem, 

\begin{definition}\label{def:BSSNWAIConstCheck} \textbf{The BSSN 
WAI Constraint Checker: }
Given an ordered action set $v = (v_i^1, \dots, v_i^j, \dots v_i^{|v|})$, where $|v|$ is the cardinality of $v$, with each action representing an expressed behaviour by the SAS, check if these observations obey the hard and soft constraints.  
\end{definition}

The second task of a WAI is defined as following:

\begin{definition}\label{def:BSSNWAITester} \textbf{The BSSN WAI Tester: }
Given a target behaviour $\hat{v} \in D(V)$ for a SAS, find the set of parameters $x$ such that if SAS is parameterised with $x$, it will produce behaviour $\hat{v}$.
\end{definition}

The first task for a WAI is theoretically not complex; purely requiring the implementation of a constraint checker; mostly with a linear complexity in the size of the constraint system. The WAI constraint checker sits within the risk assessment component of BSSN.

\begin{figure}[hptb]
\centering 
 \includegraphics[width=0.95\columnwidth]{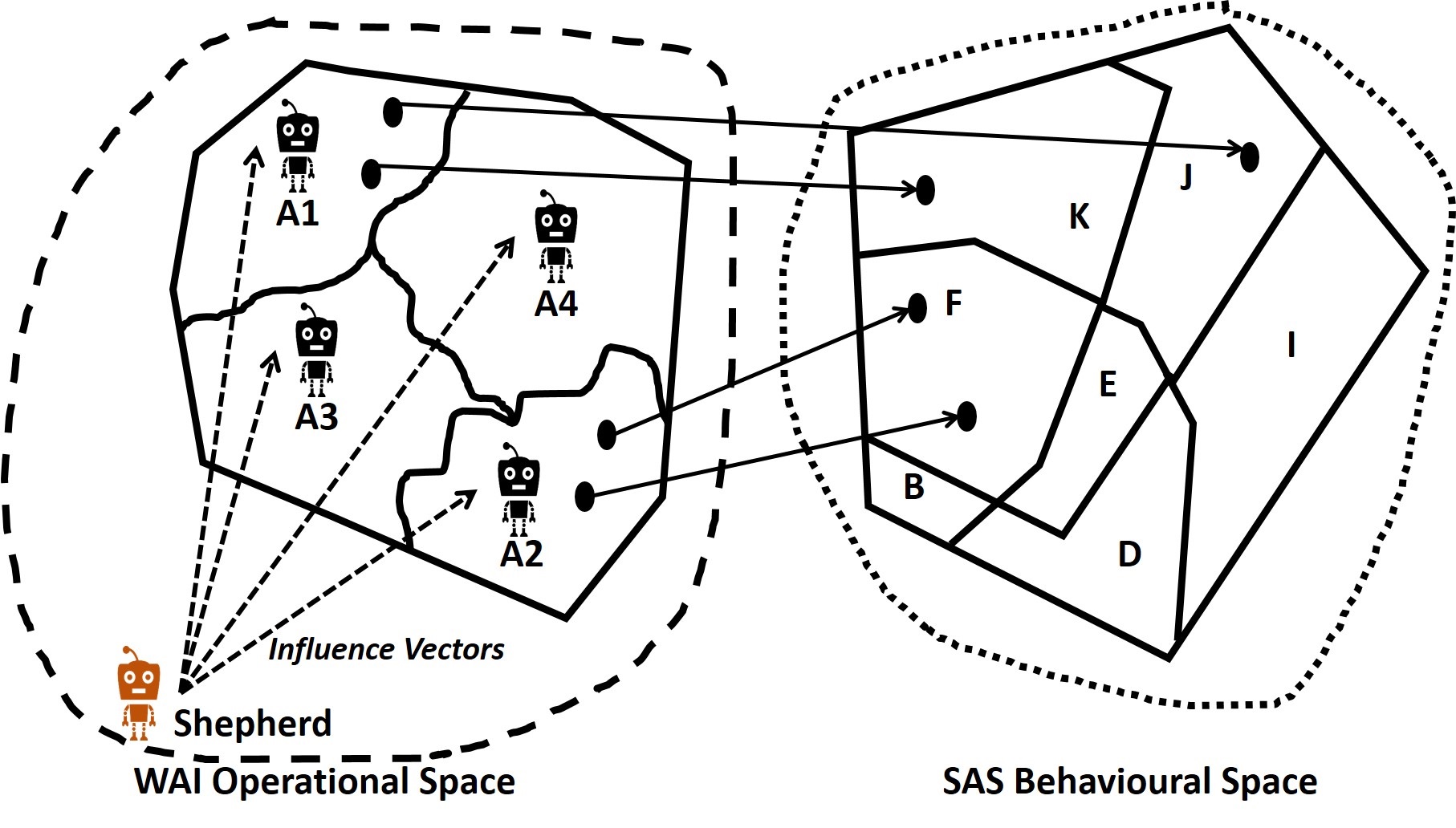}
 \caption{The mapping between the parameter space (on left) of an agent\textquoteright s sensorial information and the behavioural space (on right) or action sets of an agent.}\label{fig:spacemappings}
\end{figure}

The WAI Tester sits within the standardised experimental context component. Its task, however, could form NP-complete or even higher computational complexity problems due to the need to solve the inverse problem; that is, to find a system parameterisation that generates a specific behaviour. The complexity of this mapping is depicted in Figure~\ref{fig:spacemappings}, where as WAI $A2$ moves from one parameterisation to another in its neighbourhood, SAS generates behaviours that fall in the same sub-space $F$; that is, all movements by $A2$ generate behaviours that abide by all hard requirements. This is not the case of WAI $A1$, where a move from one parameterisation to another could generate movements from subspace $K$ to subspace $J$; that is, the move causes the SAS to shift from a behavioural sub-space that violates hard constraints but does not violate soft constraints, to a behavioural sub-space that violates both hard and soft constraints. The multimodal nature of the fitness landscape that a SAS operates upon creates ruggedness and possible discontinuities that the inverse problem may not have a solution or at best, finding one forms an NP-complete problem. 

Thus, the autonomous standardised experimental context component could form a bottleneck in slowing the system down. However, the scenarios generated from this component do not impact the operations of the system per se.

Because of the complexity imposed by the inverse problem, the set of BSSN WAI agents need to work together as a swarm; that is, a set of distributed testing agents that self-synchronise their action sets to map out the testing space. Moreover, we centralise the guidance of these BSSN WAI swarm in a Shepherd~\cite{strombom2014solving} WAI that we denote to as the $\beta$ agent. The responsibility of the shepherd WAI is to offer appropriate guidance to every $\pi_i$ to discover the different sub-spaces composing an overall behavioural space of an agent such as $S_b$.

The $\beta$ agent needs to have sufficient complexity to learn the mapping from the parameterisation space to the behavioural space of a SAS so that it is able to guide each $\pi_i$ agent. In effect, the $\beta$ agent needs to decide on the level of force it will exert on each $\pi_i$ agent so that the $\pi_i$ agent moves in the direction and speed representing the reaction vector that corresponds to the exerted force.

The formulation above mimics a sheepdog shepherding a set of sheep except in two perspectives. The first is that both the sheep and sheepdog, the $\pi_i$ and $\beta$ agents respectively, are smart AI systems. The second is that the $\beta$ agent modulates the force vector for each sheep differently, while in the biological problem, the shepherd chooses a position which generates the force vectors impacting a cluster of sheep in case of driving behaviours and a single sheep in case of collecting behaviours. As such, the position of the biological shepherd causes the set of force vectors it exerts on the sheep to be tightly coupled, while the shepherd WAI exerts force vectors that could be independent of each other if the spaces BSSN WAIs are operating on are non-overlapping.

\begin{figure}[hptb]
\centering 
 \includegraphics[width=0.95\columnwidth]{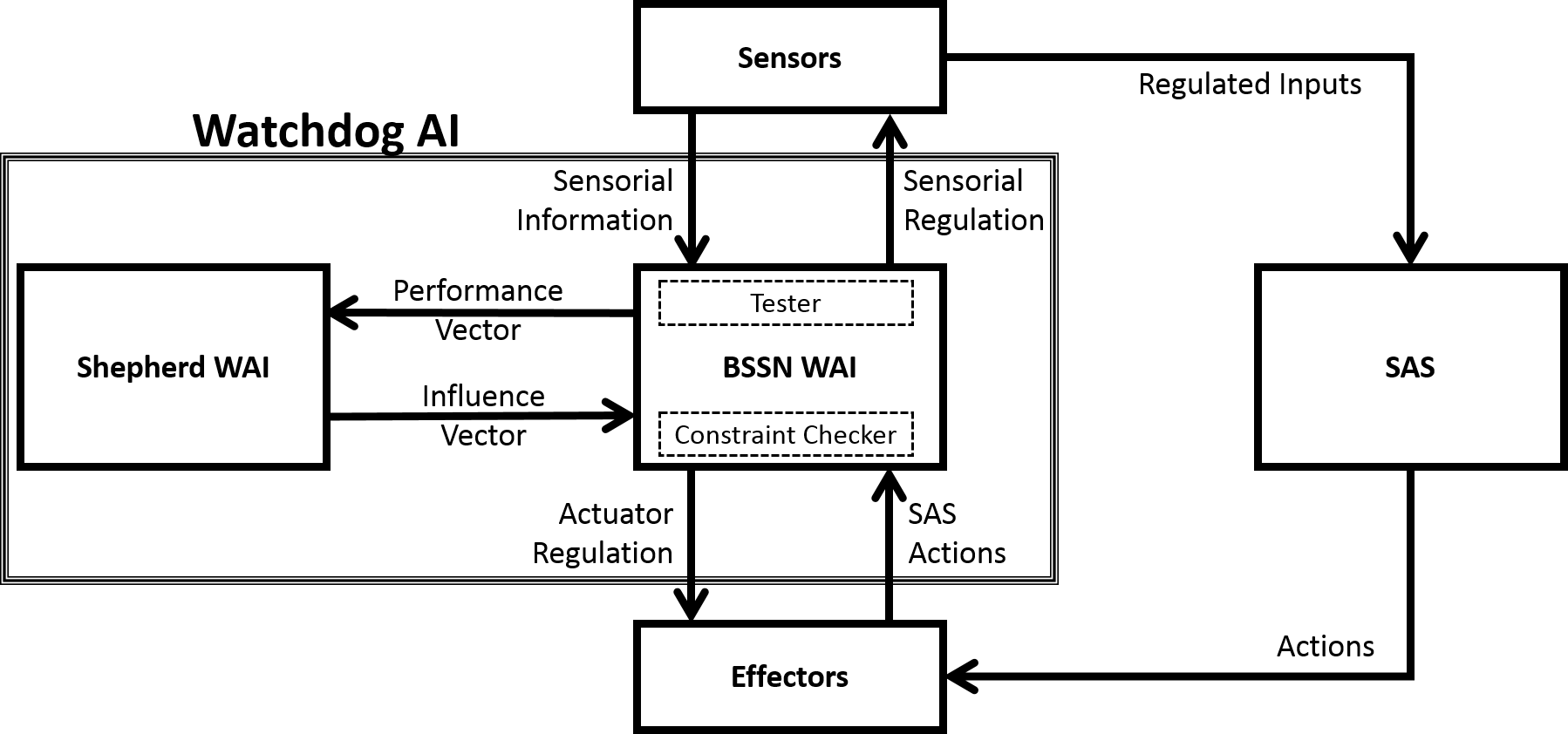}
 \caption{The watchdog AI system.}\label{fig:WAI}
\end{figure}

\section{Watchdog AI - Specifications}

In this section, we will articulate in more details the  specifications of the oracles that need to sit at the core of the WAI design. These oracles are presented in Figure~\ref{fig:WAI}. In principle, a BSSN WAI has two main components: the constraint checker to decide if a behaviour is acceptable or not and if not, how much it is violating the constraints, and the BSSN WAI tester which is responsible for standardised experimental contexts. The Shepherd WAI is responsible for evaluating the performance of each BSSN WAI agent and influence their behaviour through an influence vector that acts on the parameter space of BSSN WAI. 

\subsection{BSSN WAI Constraint Checker}

Definition~\ref{def:BSSNWAIConstCheck} provided a pragmatic description of the constraint checker. To implement this component efficiently, we need to anticipate that the specifications (eg. technological, performance, behavioural, ethical, and legal) for a SAS behaviour could form different constraint classes. Each class requires separate implementations to ensure efficiency in constraint handling.

Let $\mathcal{C} = \mathcal{C}_g \wedge \mathcal{C}_h$ be the set of all constraints in the system; that is, $C$ includes all hard ($\mathcal{C}_g = \{ g(x)\le0 \}$) and soft ($\mathcal{C}_h = \{ h(x)\le0$ \}) constraints in the system. $\mathcal{C}$ is defined over a set of variables $\mathcal{V} \in D(\mathcal{V})$, where $D$ is the domain of the variables. We distinguish four constraint classes: $\mathcal{C}^{SAT}$, $\mathcal{C}^{FD}$, $\mathcal{C}^{LR}$, and $\mathcal{C}^{NL}$ for binary, finite domains (excluding binary), linear over real numbers, and non-linear constraints, respectively. 

The notations used in the rest of the paper are: $S_v^\mathcal{V}$ denotes an ordered instantiation of $\mathcal{V}$ using $v$, 
$\vdash$ for \textquoteleft derive\textquoteright ,
$\nvdash$ for negated $\vdash$, $!$ or $\neg$ for negation, $'$ for complement, and $\bot$ for \textquoteleft falsification\textquoteright .

We use $\Psi$ to define the cost function for constraint violation. For example, $\Psi(\mathcal{C}_s^{SAT}(S_v^\mathcal{V}))$ represents the cost of constraint violation caused by substitution $v$ for propositional constraints. Two specifications below help to categorise the SAS behavioural spaces that a BSSN WAI agent needs to identify.

\begin{specification}\label{def:permissibleaction} \textbf{Permissible Action: }
An action ($S_v^\mathcal{V}$) is permissible $Perm(S_v^\mathcal{V})$ iff 
$\mathcal{C}_g^{SAT}(S_v^\mathcal{V}) \wedge \mathcal{C}_g^{FD}(S_v^\mathcal{V}) \wedge \mathcal{C}_g^{LR}(S_v^\mathcal{V}) \wedge \mathcal{C}_g^{NL}(S_v^\mathcal{V}) \nvdash \bot$. 
\end{specification}

\begin{specification}\label{def:actioninefficiencylevel} \textbf{Action Inefficiency Level: }
The level of inefficiency for an action $\Psi(S_v^\mathcal{V})$ is measured by $\Psi(S_v^\mathcal{V}) = \Psi(\mathcal{C}_s^{SAT}(S_v^\mathcal{V})) + \Psi(\mathcal{C}_s^{FD}(S_v^\mathcal{V})) + \Psi(\mathcal{C}_s^{LR}(S_v^\mathcal{V})) + \Psi(\mathcal{C}_s^{NL}(S_v^\mathcal{V}))$.
\end{specification}

We categorise SAS actions into three classes: effective and efficient permissible actions $S^{HS}$ that satisfy all hard and soft constraints, effective but inefficient permissible actions $S^{HS'}$ that satisfy all hard constraints but not all soft ones, and unpermissible actions $S^{H'}$ that violates some hard constraints. 

Efficiency in computational decision making relies on the use of an appropriate constraint handler for each specific class of constraints~\cite{Abba4}. Each constraint class is handled by a different type of constraint handler: $\mathcal{C}^{SAT}$ is handled by a satisfiability engine, $\mathcal{C}^{FD}$ by CLP(FD)~\cite{Old1}, $\mathcal{C}^{LR}$ by CLP(R)~\cite{jaffar1992clp}, and $\mathcal{C}^{NL}$ is handled by a genetic algorithm. While CLP(FD) could handle $C^{SAT}$, we prefer to split that class into an independent category because of the existence of more efficient SAT solvers today.

\begin{specification}\label{def:effectiveefficientaction} \textbf{Effective and Efficient Action: }
An action $S_v^\mathcal{V}$ is considered as effective and efficient permissible action $S^{HS}(S_v^\mathcal{V})$ when it is permissible and efficient; that is, $S^{HS}(S_v^\mathcal{V}) = Perm(S_v^\mathcal{V}) \wedge (\Psi(S_v^\mathcal{V})=0)$.
\end{specification}

\begin{specification}\label{def:effectiveinefficientaction} \textbf{Effective and Inefficient Action: }
An action $S_v^\mathcal{V}$ is considered as effective but inefficient permissible action $S^{HS'}(S_v^\mathcal{V})$ when it is permissible and efficient; that is, $S^{HS'}(S_v^\mathcal{V}) = Perm(S_v^\mathcal{V}) \wedge (\Psi(S_v^\mathcal{V}) \neq 0)$.
\end{specification}

\begin{specification}\label{def:unpermissibleaction} \textbf{Unpermissible Action: }
An action $S_v^\mathcal{V}$ is unpermissible $S^{H'}(S_v^\mathcal{V})$ when it is not permissible that is, $S^{H'}(S_v^\mathcal{V}) = \ !Perm(S_v^\mathcal{V})$.
\end{specification}

The task of the BSSN WAI Constraint Checker is to categorise an action to one of the three categories in Specifications~\ref{def:effectiveefficientaction}, \ref{def:effectiveinefficientaction}, and \ref{def:unpermissibleaction}.

\subsection{BSSN WAI Tester}

The BSSN WAI Tester was introduced in Definition~\ref{def:BSSNWAITester}. The tester does not assume that SAS acts according to a functional mapping due to the fact that the WAI agent does not have access to the internal states and/or memory of the SAS. Thus, the same input sequence could generate dramatically different output sequences. Therefore, the BSSN WAI agent does not use the input to the SAS to decide on whether the output is right or not. Instead, it relies on the output of the SAS, $S_v^\mathcal{V}$, and its own evaluation of the context to judge on the appropriateness of the output.

The tester, however, needs to be able to generate the input sequence and contextual information for continuous testing of SAS. To achieve this functionality, the tester needs an ability to learn which input sequences, $S_x^X$, could cause SAS to generate unpermissible actions (estimating SAS Failure), and which sub-space of permissible actions the SAS is able to operate (estimating SAS level of automation). Thus, four primary operators need to be performed by the tester: compression, partition, inversion, and adaptation.

Let $X^t = < x_1^t, \dots, x_M^t >$ be a sequence of possible sensorial inputs of length $M$ for a SAS at time $t$, and $V^t = < v_1^t, \dots, v_M^t >$ is the corresponding sequence of action sets that the SAS has generated. Let $p^t$ be the set of parameters used by the operators of a BSSN at a time $t$.

\begin{specification}\label{def:CompressionOperator1} \textbf{Compression Operator 1: }
Given a set of effective and efficient actions 
$S^{HS}(S_{x_s}^\mathcal{V})$, effective and inefficient actions
$S^{HS'}(S_{x_{s'}}^\mathcal{V})$, and unpermissible actions
$S^{H'}(S_{x_h}^\mathcal{V})$, where $V^t = \bigcup(x_s,x_{s'},{x_h})$, find the minimum constraint set $\mathcal{C_x}_s^t$, $\mathcal{C_x}_{s'}^t$, and $\mathcal{C_x}_h^t$ that encapsulates the three sets, respectively.
\end{specification}

\begin{specification}\label{def:CompressionOperator2} \textbf{Compression Operator 2: }
Given a set of effective and efficient actions 
$S^{HS}(S_{v_s}^\mathcal{V})$, effective and inefficient actions
$S^{HS'}(S_{v_{s'}}^\mathcal{V})$, and unpermissible actions
$S^{H'}(S_{v_h}^\mathcal{V})$, where $V^t = \bigcup(v_s,v_{s'},{v_h})$, find the minimum constraint set $\mathcal{C_v}_s^t$, $\mathcal{C_v}_{s'}^t$, and $\mathcal{C_v}_h^t$ that encapsulates the three sets, respectively.
\end{specification}

The first compression operator learns the individual clusters of the three classes of actions generated by a sequence of inputs at time $t$, while the second operator learns the corresponding clusters in the behaviour/output/action space. These clusters may contain data points that do not belong to the cluster\textquoteright s label. The partition operator fixes this by splitting and shrinking the clusters until all points in a cluster belong to the same class label with confidence level $\epsilon$.

\begin{specification}\label{def:PartitionOperator} \textbf{Partition Operator: }
Let $\mathcal{CC}^t$ be a cluster representing a class label $\mathcal{L}^t$. The cluster $\mathcal{CC}^t$ could be a cluster from Compression Operator 1 or 2; that is, it could represent $\mathcal{C_x}_s^t$, $\mathcal{C_x}_{s'}^t$, $\mathcal{C_x}_h^t$,
$\mathcal{C_v}_s^t$, $\mathcal{C_v}_{s'}^t$, or $\mathcal{C_v}_h^t$. While the confidence level on any sub-cluster is greater than $\epsilon$, find a new subspace in $\mathcal{CC}^t$ where the class label $\neg \mathcal{L}^t$. 
\end{specification}

The partition operator is computationally expensive; simply because unless the mapping $x \rightarrow v$ is linear where interval propagation methods such as CLP(BNR)~\cite{Old1} could be used, the problem is undecidable. Therefore, it is important to weight the risk of having unpermissible solution in a permissible space much higher than the risk of having a permissible solution in an unpermissible space. 

When the partition operator works on $\mathcal{C_x}_s^t$, $\mathcal{C_x}_{s'}^t$, or $\mathcal{C_x}_h^t$, it relies on the forward problem where $x$ is used as an input for SAS then the constraint checker is used to evaluate and label the output accordingly. However, when the operator works on $\mathcal{C_v}_s^t$, $\mathcal{C_v}_{s'}^t$, or $\mathcal{C_v}_h^t$, it needs to solve the inverse problem. This is where the inversion operator is called.

\begin{specification}\label{def:InversionOperator} \textbf{Inversion Operator: }
Let $v \in \mathcal{v}$, where $\mathcal{v}$ is $\mathcal{C_v}_s^t$, $\mathcal{C_v}_{s'}^t$, or $\mathcal{C_v}_h^t$. Find $x$ as an input to SAS that will generate $v$ as the corresponding action.
\end{specification}

The problem the inversion operator is working on is undecidable in the general case. It could require a sophisticated machine learning guided optimisation algorithms to find an appropriate value for $x$.

The partition operator is working on clusters generated at a particular point of time after a test sequence. After each round of testing, the old clusters need to be adapted with the new ones. This is what the adaptation operator does.

\begin{specification}\label{def:AdaptationOperator} \textbf{Adaptation Operator: }
Let $\mathcal{CC}^t$ be a cluster representing a class label $\mathcal{L}$. The cluster $\mathcal{CC}$ could be a cluster from Compression Operator 1 or 2; that is, it could represent $\mathcal{C_x}_s^t$, $\mathcal{C_x}_{s'}^t$, $\mathcal{C_x}_h^t$,
$\mathcal{C_v}_s^t$, $\mathcal{C_v}_{s'}^t$, or $\mathcal{C_v}_h^t$. Similarly define $\mathcal{CC}^t-1$ to be the equivalent cluster resultant at $t-1$. If a $\mathcal{CC}^t-1$ could be merged with any cluster $\mathcal{CC}^t$ without violating the confidence level $\epsilon$, merge the two clusters, else add $\mathcal{CC}^t-1$ to the list of clusters at time $t$.
\end{specification}

\subsection{Shepherd WAI}

The shepherd WAI operates on the ordered set of parameters, $p^t$, used by the BSSN WAI at time $t$ based on a set of performance indicators $\mathcal{I}^t$ of the system. The shepherd WAI has one operator: the influence operator. 

\begin{specification}\label{def:InfluenceOperator} \textbf{Influence Operator: }
Given $p^t$ and $\mathcal{I}^t$, find $p^{t+1}$ to improve the effectiveness and efficiency of BSSN WAIs.
\end{specification}

The influence operator may require sophisticated algorithms to adapt the parameters of BSSN WAI based on their individual performance. It may need to increase $\epsilon$ for example at the start then decay its value as BSSN explores more behavioural spaces for a SAS. It may equally increase $\epsilon$ if the partition or inverse operators are consuming significant computational costs. The detailed design of the shepherd WAI is outside the scope of this paper.

\section{Discussion}\label{Discussion}

SAS used in safety-critical systems will unlikely operate in the absence of some human involvement, be it at a low tele-operation level, mid-way at shared control level, or higher up on a supervisor-control level. WAI will operate around SAS, whether they are humans, machines or a mix. WAI could be wrapped around these systems, wrapped around a human controlling a swarm in a tele-operation scenario to ensure that the human is doing the right thing, and equally wrapped around the swarm in a supervisory control scenario or around the human-swarm system in a shared-control scenario. The design is independent of the nature of the SAS.

Our proposed WAI agents depart from current schools of thinking in autonomous systems in two ways. First, we do not use any knowledge of the internal working of SAS in the design of BSSN. As such, WAI generalises the design and avoids tight coupling and internal interdependencies within SAS that cause hidden risks. Pragmatically speaking we do not need to access other people\textquoteright s `brains' to know if their actions are right or wrong. To understand the reason motivating an action, we can either ask the person (send input queries and wait for their explanation) or develop our own `risk-balanced' portfolio of hypotheses, evaluate the subject of interest against these hypotheses, and collect sufficient evidences to select one of the hypothesis as an appropriate explanation of their action. Both approaches are fulfilled through the adoption of standardised experimental contexts. Here lies the second departure from existing literature: BSSN WAI are not static safety nets, they are adaptive and learn as guided by the shepherd WAI, but conservatively, and are designed for verifiability.

The effectiveness of the proposed BSSN WAI, contrasted to current systems with no BSSN, stems from their lifelong behavioural testing abilities to monitor and mitigate risks against the continuous learning and evolution of an autonomous system. The resultant technology aims to protect against the unpredictability of systems whom specifications are, naturally and unavoidably, incomplete at the design stage, and will remain partially known as they continue to learn, evolve and adapt when faced with new challenging contexts. BSSN WAI are designed as testing agents that act autonomously and independently over an autonomous system, are capable of regulating the autonomous system, and even have the capacity to shut it down if its behaviour gets out of control.

BSSN WAI watch the decisions of SAS to evaluate their system impact. Whether SAS is equipped with capabilities to evaluate the ethical consequences of its decisions or not, BSSN WAI agents are; therefore, the overall system acts ethically. The decoupling of WAI from SAS will mean that the pressures on industry to get SAS out is decoupled from the design and production of WAI. Designers of WAI can focus on evaluating SAS decisions on different dimensions of trust including safety and ethics, while SAS designers can focus on innovative solutions for learning and evolving the intelligence of SAS while feeling assured that WAI will be the shield that won\textquoteright t allow a decision with negative consequences to leave.

A number of challenges exist to implement WAI. The first relates to the time needed by WAIs to learn about the specific SAS that joins the system; especially that SAS is a blackbox for WAI and therefore, WAI needs to interrogate it to establish some initial bounds on the testing hypothesis space to operate from. During that time, WAI will either severely limit the performance of the SAS until it is able to estimate its behavioural spaces, or it will need to maximise its standardised experimental contexts protocols to stress test SAS rapidly enough to start opening the gates that allow the SAS to operate and actuate on the environment.

The second challenge relates to the WAIs ability to estimate the consequence of a long chain of benign actions that their compound effect becomes malignant. WAI needs to monitor both individual members of the team acting in isolation and their aggregate set of actions. This will require WAI agents to work extensively to aggregate actions in different directions to estimate aggregate consequences, which could cause a combinatorial explosion in the search space, causing the WAI agents to be overwhelmed with estimating these consequences. It could lead to a multi arm bandit game between the WAI agents and malicious SAS. There is not currently an easy fix for this challenge except that in some contexts it will be less of a problem than some other contexts. Having a human in supervisory control role of the shepherd WAI to oversee the decisions of BSSN WAI is maybe unavoidable in certain context. However, it will still bring other challenges including the mismatch between the speed of processing of a human when compared to the speed of processing of well-resourced computational WAI.

The discussion of WAI agents so far relied on a simple design working on a level of abstraction that allows the problem to be represented mathematically. The complexity of WAI agents, however, pose another challenge of how to engineer the architecture of BSSN and Shepherd WAI to ensure that the system will scale while it has the right level of complexity to manage the behaviour space of the SAS agents it is responsible for? A hierarchical organisation of BSSN and Shepherd WAI agents may be needed in complex applications. 

\section{Conclusion}\label{Conclusion}

We proposed the design of artificial intelligence (AI) agents to act as watchdogs for other AI agents and smart autonomous systems. The Watchdog AI (WAI) concept is presented along with a discussion of the requirements and design principles for these watchdogs to operate. The design of the WAI agents was inspired by the biological phenomena of sheepdogs shepherding a group of sheep. The challenges associated with the implementation of WAI agents were discussed. While there are challenges, the motivation behind the WAI concept and benefits of designing WAI were established to demonstrate that the concept of WAI might be unavoidable; especially in the use of AI in safety-critical systems.



\end{document}